\def\year{2020}\relax
\documentclass[letterpaper]{article} 
\usepackage{aaai20}  
\usepackage{times}  
\usepackage{helvet} 
\usepackage{courier}  
\usepackage[hyphens]{url}  
\usepackage{graphicx} 
\usepackage{graphicx}  
\usepackage{comment}
\usepackage{rotating}
\usepackage{amsmath}
\usepackage{amsthm}
\usepackage{amssymb}
\usepackage{graphics}
\usepackage{transparent}
\usepackage{multirow}
\usepackage{algorithm}
\usepackage[noend]{algpseudocode}
\usepackage[Symbolsmallscale]{upgreek}
\usepackage{centernot}
\usepackage{enumitem}

\algnewcommand{\LineComment}[1]{\State \(\triangleright\) #1}

\newcommand{\T}{\intercal}

\newcommand{\bX}{ \mbox{\bf X}}

\newtheorem*{thm*}{Theorem}

\theoremstyle{definition}

\title{Sex Trafficking Detection with Ordinal Regression Neural Networks}
\author{Longshaokan Wang\textsuperscript{\rm 1}\thanks{This work was done when Wang was a PhD student at North Carolina State University.}, Eric Laber\textsuperscript{\rm 2}, Yeng Saanchi\textsuperscript{\rm 3}, Sherrie Caltagirone\textsuperscript{\rm 4} \\ 
\textsuperscript{\rm 1}Alexa AI, Amazon, \textsuperscript{\rm 23}Department of Statistics, North Carolina State University, \textsuperscript{\rm 4}Global Emancipation Network \\ 
\textsuperscript{\rm 1}longsha@amazon.com, \textsuperscript{\rm 23}\{eblaber, ysaanch\}@ncsu.edu, \textsuperscript{\rm 4}sherrie@globalemancipation.ngo 
}

\begin{document}

\maketitle

\begin{abstract}
Sex trafficking is a global epidemic. Escort websites are a primary
vehicle for selling the services of such trafficking victims and
thus a major driver of trafficker revenue. Many law enforcement
agencies do not have the resources to manually identify leads from
the millions of escort ads posted across dozens of public websites.
We propose an ordinal regression neural network to identify escort
ads that are likely linked to sex trafficking. Our model uses a
modified cost function to mitigate inconsistencies in predictions
often associated with nonparametric ordinal regression and leverages
recent advancements in deep learning to improve prediction accuracy.
The proposed method significantly improves on the
previous state-of-the-art on Trafficking-10K, an
expert-annotated dataset of escort ads. Additionally, because
traffickers use acronyms, deliberate typographical errors, and
emojis to replace explicit keywords, we demonstrate how to expand
the lexicon of trafficking flags through word embeddings and t-SNE.
\end{abstract}

\section{Introduction}
\label{sec:intro}

Globally, human trafficking is one of the fastest growing
crimes and, with annual profits estimated to be in excess
of 150 billion USD, it is also among the most lucrative 
\cite{AminS2010}. Sex trafficking is a form of human trafficking
which involves sexual exploitation through 
coercion. Recent estimates suggest that nearly 
4 million adults 
and 1 million children are being victimized globally on any
given day; furthermore, it is estimated that 99 percent of
victims are  female \cite{ILO2017}. Escort websites
are an increasingly popular vehicle for selling the services of
trafficking victims. According to a recent survivor survey
\cite{THORN2018}, 38\% of underage trafficking victims who were
enslaved prior to 2004 were advertised online, and that number rose to
75\% for those enslaved after 2004.
Prior to its shutdown in April 2018, the website Backpage was
the most frequently used online advertising
platform; other popular websites used for advertising escort service include
Craigslist, Redbook, SugarDaddy, and Facebook
\cite{THORN2018}.  Despite the seizure of Backpage,
there were nearly 150,000 new online sex advertisements
posted {\em per day} in the U.S. alone in late 2018 \cite{TarinelliR2018}; 
even with many of these new
ads being re-posts of existing ads and traffickers often posting
multiple ads for the same victims \cite{THORN2018}, this volume is staggering.

Because of their ubiquity and public access, escort websites are
a rich resource for anti-trafficking
operations.  However, many law enforcement agencies do not have the
resources to sift through the volume of escort ads to identify
those coming from potential traffickers. One scalable and efficient
solution is to build a statistical model to predict the likelihood of
an ad coming from a trafficker using a dataset annotated by
anti-trafficking experts.
We propose an ordinal regression neural network tailored for text
input. This model comprises three components: (i) a Word2Vec model
\cite{MikolovT2013b} that maps each word from the text input to a
numeric vector, (ii) a gated-feedback recurrent neural network
\cite{ChungJ2015} that sequentially processes the word vectors,
and (iii) an ordinal regression layer \cite{ChengJ2008} that
produces a predicted ordinal label. We use a modified cost
function to mitigate inconsistencies in predictions associated with
nonparametric ordinal regression.  We also leverage several
regularization techniques for deep neural networks to further improve
model performance, such as residual connection \cite{HeK2016} and
batch normalization \cite{IoffeS2015}. We conduct our experiments
on Trafficking-10k \cite{TongE2017}, a dataset of escort ads for
which anti-trafficking experts assigned each sample one of seven
ordered labels ranging from ``1: Very Unlikely (to come from
traffickers)'' to ``7: Very Likely''. Our proposed model significantly
outperforms previously published models \cite{TongE2017} on
Trafficking-10k as well as a variety of baseline ordinal regression
models. In addition, we analyze the emojis
used in escort ads with Word2Vec and
t-SNE \cite{MaatenL2008}, and we show that the lexicon of
trafficking-related emojis can be subsequently expanded.

The main contributions of this paper are summarized as follows: 
1. We propose a neural network architecture for text data with ordinal labels, 
which outperforms the previous state-of-the-art \cite{TongE2017} on Trafficking-10k.
2. We propose a simple penalty term in the cost function 
to mitigate the monotonicity violation and 
improve the interpretability of the output of the ordinal regression layer, 
where the monotonicity violation was previously deemed too computationally costly to resolve \cite{NiuZ2016}.
3. We provide a qualitative analysis on the top escort ads flagged by our model, 
which offers patterns that anti-trafficking experts can potentially confirm or make use of.
4. We provide an emoji analysis that shows how to use unsupervised learning techniques 
on the raw data to generate leads on new trafficking key words.
5. We open source our code base and trained model to encourage further research on trafficking detection 
and to allow the law enforcement to make use of our research for free.

In Section \ref{sec:rel}, we discuss related work on human
trafficking detection and ordinal regression.  In Section
\ref{sec:method}, we present our proposed model and detail its 
components.  In Section \ref{sec:exp}, we present the
experimental results, including the Trafficking-10K benchmark,
a qualitative analysis of the predictions on raw
data, and the emoji analysis.
In Section \ref{sec:dis}, we summarize
our findings and discuss future work.

\section{Related Work}
\label{sec:rel}
\textbf{Trafficking detection:}  There have been several
software products designed to aid anti-trafficking
efforts.  
Examples include 
Memex\footnote{darpa.mil/program/memex} which focuses on search
functionalities in the dark web; Spotlight\footnote{htspotlight.com}
which flags suspicious ads
and links images appearing in multiple ads; Traffic
Jam\footnote{marinusanalytics.com/trafficjam} which
seeks to identify patterns that 
connect multiple ads to the same trafficking organization;
and TraffickCam\footnote{traffickcam.com} which aims to construct a crowd-sourced 
database of hotel room images to geo-locate victims.
These research efforts have largely been isolated, and 
few research articles on machine learning for
trafficking detection have been published.
Closest to our work is the Human Trafficking Deep Network (HTDN) \cite{TongE2017}. HTDN has three main
components: a language network that uses pretrained word embeddings
and a long short-term memory network (LSTM) to
process text input; a vision network that uses a convolutional network
to process image input; and another convolutional network to combine
the output of the previous two networks and produce a binary
classification. Compared to the language network in HTDN, our model
replaces LSTM with a gated-feedback recurrent neural network, adopts
certain regularizations, and uses an ordinal regression layer on top.
It significantly improves HTDN's benchmark despite only using text
input.  As in the work of \citeauthor{TongE2017} \shortcite{TongE2017}, 
we pre-train word embeddings using a skip-gram model 
\cite{MikolovT2013b}  applied to unlabeled data from
escort ads, however, we go further by analyzing the
emojis' embeddings and thereby expand the trafficking lexicon.

\textbf{Ordinal regression:}
We briefly review ordinal regression before introducing the
proposed methodology.  
We assume that the training data are 
$D_\text{train} = \left\{\left(\bX_i, Y_i\right)\right\}_{i=1}^n$,
where $\bX_i\in\mathcal{X}$ are the features and
$Y_i\in\mathcal{Y}$ is the response;
$\mathcal{Y}$ is the set 
of $k$ ordered labels $\{1, 2, \ldots, k\}$ with 
$1 \prec 2 \ldots \prec k$. Many ordinal regression methods 
learn a composite map $\eta = h \circ g$, where
$g: \mathcal{X} \rightarrow \mathbb{R}$ and
$h: \mathbb{R} \rightarrow \{1, 2, \ldots, k\}$ have
the interpretation that $g(\bX)$ is a latent ``score''
which is subsequently discretized into a category by $h$.
$\eta$ is often estimated by
empirical risk minimization, i.e., 
by minimizing a 
loss function $C\{\eta(\bX), Y\}$ averaged over the training data.
Standard choices of $\eta$ and $C$ 
are reviewed by \citeauthor{RennieJ2005} \shortcite{RennieJ2005}.

Another common approach to ordinal regression, which we
adopt in our proposed method, is to transform the label prediction into
a series of $k - 1$ binary classification sub-problems, wherein the $i$th sub-problem 
is to predict whether the true label exceeds 
$i$ (\citeauthor{FrankE2001} \citeyear{FrankE2001}; \citeauthor{LiL2006} \citeyear{LiL2006}).  
For example, one might use a series
of logistic regression models to estimate the conditional
probabilities
$f_i(\mathbf{X}) = P(Y > i \big|\mathbf{X})$ for each $i=1,\ldots,k-1$.
\citeauthor{ChengJ2008} (\citeyear{ChengJ2008}) estimated
these probabilities jointly using a neural network;
this was later extended to image data \cite{NiuZ2016}
as well as text data (\citeauthor{IrsoyO2015} \citeyear{IrsoyO2015}; \citeauthor{RuderS2016} \citeyear{RuderS2016}).  
However, as acknowledged by \citeauthor{ChengJ2008} (\citeyear{ChengJ2008}), the estimated probabilities
need not respect the ordering
$f_i(\mathbf{X}) \ge f_{i+1}(\mathbf{X})$ for all $i$ and
$\mathbf{X}$.  We force our estimator to respect this
ordering through a penalty on its violation.  

\section{Method}
\label{sec:method}

\begin{figure}[t]
\centering
\includegraphics[width=1.0\columnwidth]{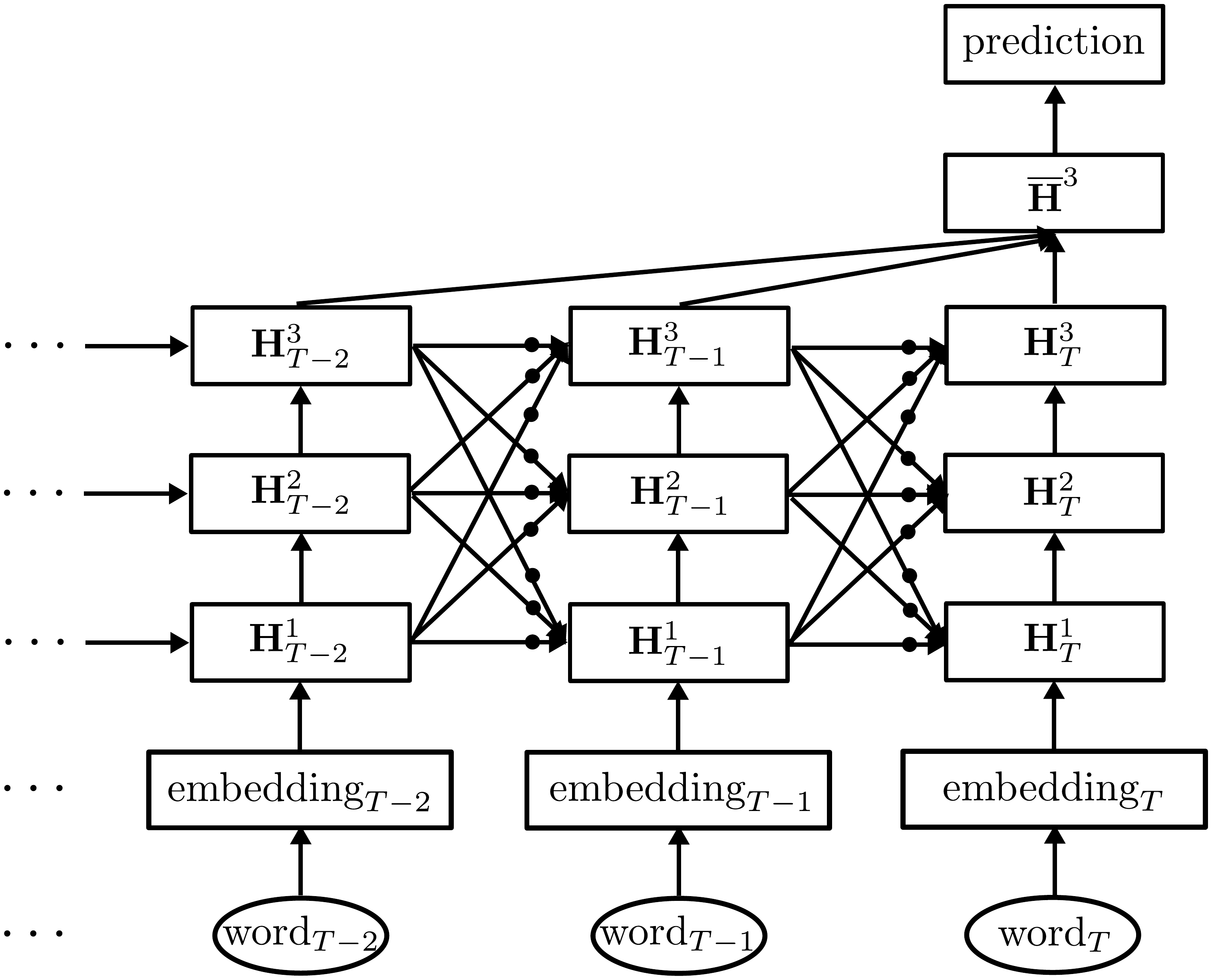}
\caption{Overview of the ordinal regression neural network for text input. $\mathbf{H}$ represents 
            a hidden state in a gated-feedback recurrent neural network.}
\label{fig-ornn}
\end{figure}

Our proposed ordinal regression model consists of the following
three components: Word embeddings pre-trained by a Skip-gram model, a
gated-feedback recurrent neural network that constructs summary
features from sentences, and a multi-labeled logistic regression layer
tailored for ordinal regression.
See Figure \ref{fig-ornn} for a schematic.  
The details of its components and
their respective alternatives are discussed below.

\subsection{Word Embeddings}
\label{sec:embedding}

Vector representations of words, also known as word embeddings, can be
obtained through unsupervised learning on a large text corpus so that
certain linguistic regularities and patterns are encoded. Compared to
Latent Semantic Analysis \cite{DumaisS2004}, embedding algorithms
using neural networks are particularly good at preserving linear
regularities among words in addition to grouping similar words
together \cite{MikolovT2013a}.
Such embeddings can in turn 
help other algorithms achieve better performances in various natural language processing tasks 
\cite{MikolovT2013b}.

Unfortunately, the escort ads contain a plethora of emojis, acronyms,
and (sometimes deliberate) typographical errors that are not
encountered in more standard text data, which suggests that it is
likely better to learn word embeddings from scratch on a large
collection of escort ads instead of using previously published
embeddings \cite{TongE2017}. We use 168,337 ads scraped from
Backpage as our training corpus and the Skip-gram model with Negative
sampling \cite{MikolovT2013b} as our model.

\subsection{Gated-Feedback Recurrent Neural Network}
\label{sec:gfrnn}

To process entire sentences and paragraphs after mapping the words to
embeddings, we need a model to handle sequential data. Recurrent
neural networks (RNNs) have recently seen great success at modeling
sequential data, especially in natural language processing tasks
\cite{LeCunY2015}. On a high level, an RNN is a neural network
that processes a sequence of inputs one at a time, taking the summary
of the sequence seen so far from the previous time point as an
additional input and producing a summary for the next time point. One
of the most widely used variations of RNNs, a Long short-term memory
network (LSTM), uses various gates to control the information flow and
is able to better preserve long-term dependencies in the running
summary compared to a basic RNN (see \citeauthor{GoodfellowI2016} \citeyear{GoodfellowI2016} and references therein).  In our implementation, we use
a further refinement of multi-layed LSTMs, Gated-feedback
recurrent neural networks (GF-RNNs), which tend
to capture dependencies across different timescales more
easily \cite{ChungJ2015}.

Regularization techniques for neural networks including Dropout
\cite{SrivastavaN2014}, Residual connection \cite{HeK2016},
and Batch normalization \cite{IoffeS2015} are added to GF-RNN for
further improvements.

After GF-RNN processes an entire escort ad, the average of the hidden
states of the last layer
becomes the input for the multi-labeled 
logistic regression layer which we discuss next.   

\subsection{Multi-Labeled Logistic Regression Layer}
\label{sec:multi-label}
As noted previously,
the ordinal regression problem can be cast into a series of binary
classification problems and thereby utilize the large repository 
of available classification
algorithms (\citeauthor{FrankE2001} \citeyear{FrankE2001}; \citeauthor{LiL2006} \citeyear{LiL2006}; \citeauthor{NiuZ2016} \citeyear{NiuZ2016}).  One formulation
is as follows. Given $k$ total ranks, the $i$-th binary classifier is
trained to predict the probability that a sample $\bX$ has rank larger
than $i: \widehat{f}_i(\bX) = \widehat{\text{P}}(Y > i \lvert \bX)$.
Then the predicted rank is
$$
\widehat{Y} = 1 + \sum_{i=1}^{k-1} \text{Round}\left\{\widehat{f}_i(\bX)\right\}.
$$

\begin{figure}[t]
\centering
\includegraphics[width=1.0\columnwidth]{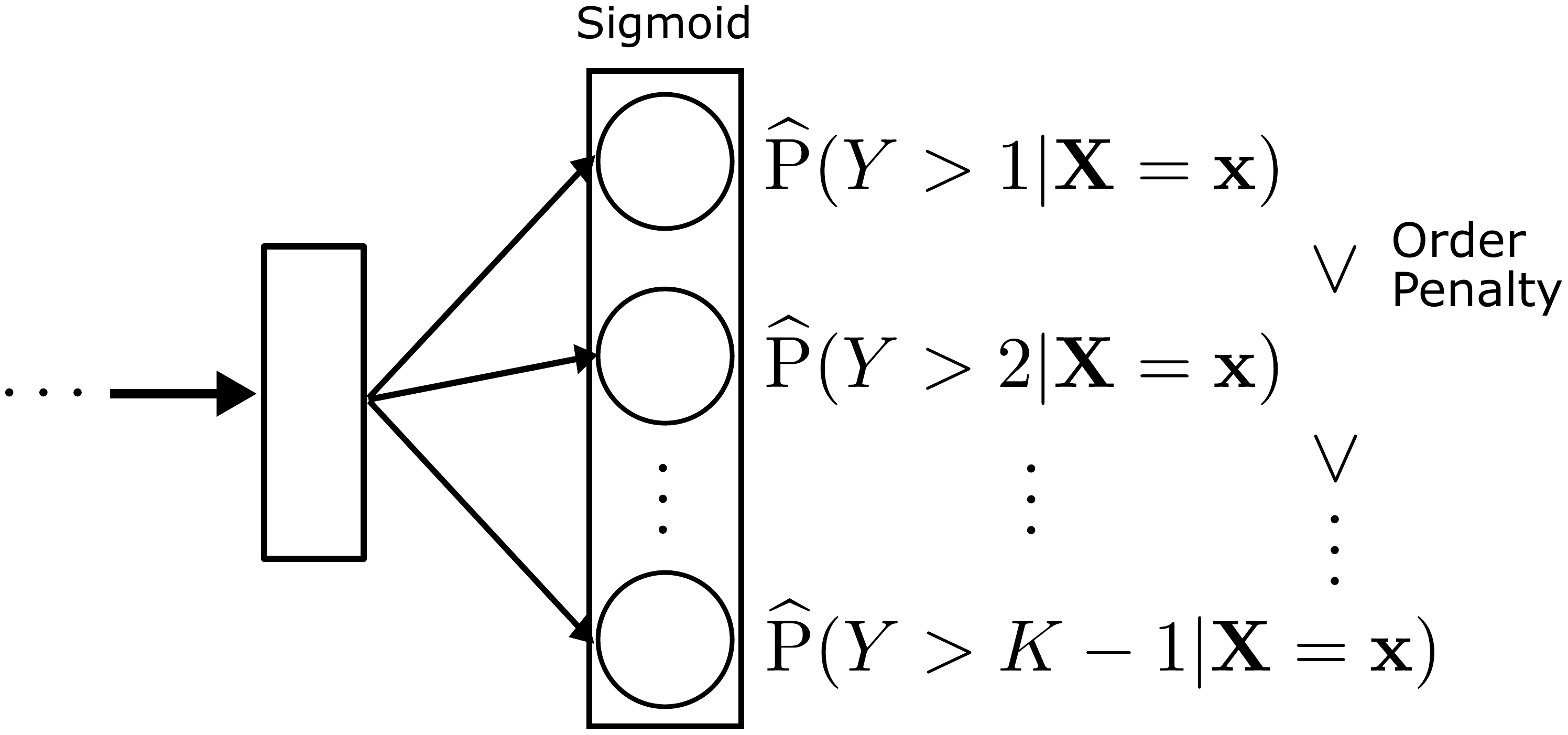}
\caption{Ordinal regression layer with order penalty.}
\label{multi_label}
\end{figure}

In a classification task, the final layer of a deep neural network is
typically a softmax layer with dimension equal
to the number of classes
\cite{GoodfellowI2016}. Using the
ordinal-regression-to-binary-classifications formulation described
above, \citeauthor{ChengJ2008}  (\citeyear{ChengJ2008}) replaced the softmax
layer in their neural network with a $(k-1)$-dimensional sigmoid
layer, where each neuron serves as a binary classifier (see Figure
\ref{multi_label} but without the order penalty to be discussed later).

With the sigmoid activation function, the output of the $i$th
neuron can be viewed as the predicted probability that the sample has
rank greater\footnote{Actually, in \citeauthor{ChengJ2008}'s original
  formulation, the final layer is $k$-dimensional with the $i$-th
  neuron predicting the probability that the sample has rank greater
  than \textsl{or equal} to $i$. This is redundant because the first
  neuron should always be equal to 1. Hence we make the slight
  adjustment of using only $k-1$ neurons.} than $i$. Alternatively, the entire
sigmoid layer can be viewed as performing multi-labeled logistic
regression, where the $i$th label is the indicator of the sample's
rank being greater than $i$. The training data are thus re-formatted
accordingly so that response variable for a sample with rank $i$ becomes
$(\mathbf{1}_{i-1}^\T, \mathbf{0}_{k-i}^\T)^\T$.  The $k-1$ binary
classifiers share the features constructed by the earlier layers of
the neural network and can be trained jointly with mean squared error
loss. A key difference between the multi-labeled logistic
regression and the naive classification (ignoring the order and
treating all ranks as separate classes) is that the loss for
$\widehat{Y} \neq Y$ is constant in the naive classification but
proportional to $\lvert \widehat{Y} - Y \lvert$ in the multi-labeled
logistic regression.  

\citeauthor{ChengJ2008}'s (\citeyear{ChengJ2008}) final layer was preceded by
a simple feed-forward network. In our case, word embeddings and GF-RNN
allow us to construct a feature vector of fixed length from text
input, so we can simply attach the multi-labeled logistic regression
layer to the output of GF-RNN to complete an ordinal regression neural
network for text input.

The violation of the
monotonicity in the estimated probabilities (e.g., $\widehat{f}_i(\bX) < \widehat{f}_{i+1}(\bX)$ for some $\bX$ and $i$) has
remained an open issue since the original ordinal regression neural network proposal of \citeauthor{ChengJ2008} (\citeyear{ChengJ2008}).
This is perhaps owed in part to the
belief that correcting this issue would significantly increase
training complexity \cite{NiuZ2016}.  
We propose an effective and computationally
efficient solution to avoid the conflicting predictions
as follows:  penalize such
conflicts in the training phase by adding
$$
P(\bX; \lambda) = \lambda \sum_{i=1}^{k-2} \max\left\{\widehat{f}_{i+1}(\bX) - \widehat{f}_i(\bX), 0\right\}
$$
to the loss function for a sample $\bX$, where $\lambda$ is a penalty
parameter (Figure \ref{multi_label}). 
For sufficiently large $\lambda$ the estimated probabilities
will respect the monotonicity condition; respecting this
condition improves the interpretability of the predictions, which is vital in applications like the one we
consider here as stakeholders are given the
estimated probabilities. 
We also hypothesize that the order penalty may serve 
as a regularizer to improve each binary classifier 
(see the ablation test in Section \ref{sec:ablation}).

All three components of our model (word embeddings, GF-RNN, and
multi-labeled logistic regression layer) can be trained jointly, with
word embeddings optionally held fixed or given a smaller learning rate
for fine-tuning. The hyperparameters for all components are given in
the Appendix. They are selected according to either literature or
grid-search.

\section{Experiments}
\label{sec:exp}

We first describe the datasets we use to train and evaluate our
models. Then we present a detailed comparison of our proposed model
with commonly used ordinal regression models as well as the
previous state-of-the-art classification model by \citeauthor{TongE2017}
(\citeyear{TongE2017}). To assess the effect of each component in our
model, we perform an ablation test where the components are swapped by
their more standard alternatives one at a time.  Next, we perform a
qualitative analysis on the model predictions on the raw data, which are scraped from a
different escort website than the one that provides the labeled
training data.  Finally, we conduct an emoji analysis using the word
embeddings trained on raw escort ads.

\subsection{Datasets}
\label{sec:data}

We use raw texts scraped from Backpage and TNABoard to pre-train the
word embeddings, and use the same labeled texts \citeauthor{TongE2017} (\citeyear{TongE2017}) used to conduct model comparisons.  The raw
text dataset consists of 44,105 ads from TNABoard and 124,220 ads from
Backpage.  Data cleaning/preprocessing includes joining the title and
the body of an ad; adding white spaces around every emoji so that it
can be tokenized properly; stripping tabs, line breaks, punctuations,
and extra white spaces; removing phone numbers; and converting all
letters to lower case. We have ensured that the raw dataset 
has no overlap with the labeled dataset to avoid bias in test
accuracy. While it is possible to scrape more raw data, we did not
observe significant improvements in model performances when the size
of raw data increased from $\sim$70,000 to $\sim$170,000, hence we
assume that the current raw dataset is sufficiently large.

The labeled dataset is called Trafficking-10k. It consists of 12,350
ads from Backpage labeled by experts in human trafficking detection\footnote{
  Backpage was seized by FBI in April 2018, but we have
  observed that escort ads across different websites are often
  similar, and a survivor survey shows that traffickers post their
  ads on multiple websites \cite{THORN2018}.  Thus, we argue that the training data
  from Backpage are still useful, which is empirically supported by
  our qualitative analysis in Section \ref{sec:qualitative}.}
\cite{TongE2017}.  Each label is one of seven ordered levels of
likelihood that the corresponding ad comes from a human
trafficker. Descriptions and sample proportions of the labels are in
Table \ref{data}. The original Trafficking-10K includes both texts and
images, but as mentioned in Section \ref{sec:intro}, only the texts
are used in our case.  We apply the same preprocessing to
Trafficking-10k as we do to raw data.
\begin{table*}[htb]
\begin{center}
\begin{tabular}{l|c|c|c|c|c|c|c}
\hline
Label & 1 & 2 & 3 & 4 & 5 & 6 & 7 \\
\hline
\multirow{2}{*}{Description} & Strongly & \multirow{2}{*}{Unlikely} & Slightly & \multirow{2}{*}{Unsure} & Weakly & \multirow{2}{*}{Likely} & Strongly \\
& Unlikely & & Unlikely & & Likely & & Likely \\
\hline
Count & 1,977 & 1,904 & 3,619 & 796 & 3,515 & 457 & 82 \\
\hline
\end{tabular}
\caption{\label{data}Description and distribution of labels in Trafficking-10K.}
\end{center}
\end{table*}

\subsection{Comparison with Baselines}

We compare our proposed ordinal regression neural network (ORNN) to
Immediate-Threshold ordinal logistic regression (IT)
\cite{RennieJ2005}, All-Threshold ordinal logistic regression
(AT) \cite{RennieJ2005}, Least Absolute Deviation (LAD) (\citeauthor{BloomfieldP1980} \citeyear{BloomfieldP1980}; \citeauthor{NarulaS1982} \citeyear{NarulaS1982}), and multi-class logistic
regression (MC) which ignores the ordering. The primary evaluation
metrics are Mean Absolute Error (MAE) and macro-averaged Mean Absolute
Error ($\text{MAE}^M$) \cite{BaccianellaS2009}.
To compare our model with the previous
state-of-the-art classification model for escort ads, the Human
Trafficking Deep Network (HTDN) \cite{TongE2017}, we also
polarize the true and predicted labels into two classes, ``1-4:
Unlikely'' and ``5-7: Likely''; then we compute the binary
classification accuracy (Acc.) as well as the weighted binary
classification accuracy (Wt. Acc.) given by
$$
\resizebox{1.0\columnwidth}{!}{$
\text{Wt. Acc.} = \frac{1}{2} \left(\frac{\text{True Positives}}{\text{Total Positives}} + \frac{\text{True Negatives}}{\text{Total Negatives}}\right).
$}
$$ 
Note that for applications in human trafficking detection, MAE and Acc. are of primary interest. 
Whereas for a more general comparison among the models, the class imbalance robust metrics, 
$\text{MAE}^M$ and Wt. Acc., might be more suitable. Bootstrapping or increasing the weight of 
samples in smaller classes can improve $\text{MAE}^M$ and Wt. Acc. at the cost of MAE and Acc..

The text data need to be vectorized before they can be fed into the baseline models (whereas 
vectorization is built into ORNN). The standard practice is to tokenize the texts using n-grams 
and then create weighted term frequency vectors using the term frequency (TF)-inverse document 
frequency (IDF) scheme (\citeauthor{BeelJ2016} \citeyear{BeelJ2016}; \citeauthor{ManningC2009} \citeyear{ManningC2009}). The specific variation we use is 
the recommended unigram + sublinear TF + smooth IDF (\citeauthor{ManningC2009} \citeyear{ManningC2009}; \citeauthor{PedregosaF2011} \citeyear{PedregosaF2011}). 
Dimension reduction techniques such as Latent Semantic Analysis 
\cite{DumaisS2004} can be optionally applied to the frequency vectors, but \citeauthor{SchullerB2015} 
(\citeyear{SchullerB2015}) concluded from their experiments that dimension reduction on frequency 
vectors actually hurts model performance, which our preliminary experiments agree with.

All models are trained and evaluated using the same (w.r.t. data shuffle and split) 10-fold 
cross-validation (CV) on Trafficking-10k, except for HTDN, whose result is read from the original 
paper \cite{TongE2017}\footnote{The authors of HTDN used a single train-validation-test 
split instead of CV.}. During each train-test split, $2/9$ of the training set is further reserved as 
the validation set for tuning hyperparameters such as L2-penalty in IT, AT and LAD, and learning 
rate in ORNN. So the overall train-validation-test ratio is 70\%-20\%-10\%. We report the mean 
metrics from the CV in Table \ref{comp}. As previous research has pointed out that there is no 
unbiased estimator of the variance of CV \cite{BengioY2004}, we report the naive 
standard error treating metrics across CV as independent. Recall that a 95\% confidence interval is roughly the point estimate $\pm \,1.96 \times \text{the standard error}$.

\begin{table*}[htb]
\begin{center}
\begin{tabular}{l|c|c|c|c}
\hline
Model & MAE & $\text{MAE}^M$ & Acc. & Wt. Acc. \\
\hline
ORNN & \textbf{0.769} (0.009) & \textbf{1.238} (0.016) & \textbf{0.818} (0.003) & \textbf{0.772} (0.004) \\
\hline
IT & 0.807 (0.010) & \textbf{1.244} (0.011) & 0.801 (0.003) & \textbf{0.781} (0.004) \\
\hline
AT & \textbf{0.778} (0.009) & 1.246 (0.012) & \textbf{0.813} (0.003) & 0.755 (0.004) \\
\hline
LAD & 0.829 (0.008) & 1.298 (0.016) & 0.786 (0.004) & 0.686 (0.003) \\
\hline
MC & 0.794 (0.012) & 1.286 (0.018) & 0.804 (0.003) & 0.767 (0.004) \\
\hline
HTDN & - & - & 0.800 & 0.753 \\
\hline
\end{tabular}
\caption{\label{comp}Comparison of the proposed ordinal regression neural network (ORNN) against 
Immediate-Threshold ordinal logistic regression (IT), All-Threshold ordinal logistic regression (AT), 
Least Absolute Deviation (LAD), multi-class logistic regression (MC), and the Human Trafficking 
Deep Network (HTDN) in terms of Mean Absolute Error (MAE), macro-averaged Mean Absolute 
Error ($\text{MAE}^M$), binary classification accuracy (Acc.) and weighted binary classification 
accuracy (Wt. Acc.). The results are averaged across 10-fold CV on Trafficking-10k with naive 
standard errors in the parentheses. The best and second best results are highlighted.}
\end{center}
\end{table*}

We can see that ORNN has the best MAE, $\text{MAE}^M$ and Acc. as well as a close 2nd 
best Wt. Acc. among all models. Its Wt. Acc. is a substantial improvement over HTDN despite 
the fact that the latter use both text and image data. It is important to note that HTDN is trained 
using binary labels, whereas the other models are trained using ordinal labels and then have 
their ordinal predictions converted to binary predictions. This is most likely the reason that 
even the baseline models except for LAD can yield better Wt. Acc. than HTDN, confirming 
our earlier claim that polarizing the ordinal labels during training may lead to information loss.

\subsection{Ablation Test}
\label{sec:ablation}

To ensure that we do not unnecessarily complicate our ORNN model, and
to assess the impact of each component on the final model performance,
we perform an ablation test. Using the same CV and evaluation
metrics, we make the following replacements separately and re-evaluate
the model: 1. Replace word embeddings pre-trained from skip-gram model
with randomly initialized word embeddings; 2. replace gated-feedback
recurrent neural network with long short-term memory network (LSTM);
3. disable batch normalization; 4. disable residual connection;
5. replace the multi-labeled logistic regression layer with a softmax
layer (i.e., let the model perform classification, treating the
ordinal response variable as a categorical variable with $k$ classes);
6. replace the multi-labeled logistic regression layer with a
1-dimensional linear layer (i.e., let the model perform regression,
treating the ordinal response variable as a continuous variable) and
round the prediction to the nearest integer during testing; 7. set the
order penalty to 0. The results are shown in Table \ref{ablation}.

\begin{table*}[htb]
\begin{center}
\begin{tabular}{l|c|c|c|c}
\hline
Model & MAE & $\text{MAE}^M$ & Acc. & Wt. Acc. \\
\hline
0. Proposed ORNN & \textbf{0.769} (0.009) & \textbf{1.238} (0.016) & \textbf{0.818} (0.003) & \textbf{0.772} (0.004) \\
\hline
1. Random Embeddings & 0.789 (0.007) & 1.254 (0.013) & 0.810 (0.002) & 0.757 (0.003) \\
\hline
2. LSTM & 0.778 (0.009) & 1.261 (0.021) & 0.815 (0.003) & 0.764 (0.003) \\
\hline
3. No Batch Norm. & 0.780 (0.009) & 1.311 (0.013) & 0.815 (0.003) & 0.754 (0.004) \\
\hline
4. No Res. Connect. & \textbf{0.775} (0.008) & 1.271 (0.020) & \textbf{0.816} (0.003) & 0.766 (0.004) \\
\hline
5. Classification & 0.785 (0.012) & 1.253 (0.017) & 0.812 (0.004) & \textbf{0.780} (0.004) \\
\hline
6. Regression & 0.850 (0.009) & 1.279 (0.016) & 0.784 (0.004) & 0.686 (0.006) \\
\hline
7. No Order Penalty & \textbf{0.769} (0.009) & \textbf{1.251} (0.016) & \textbf{0.818} (0.003) & 0.769 (0.004) \\
\hline
\end{tabular}
\caption{\label{ablation}Ablation test. Except for models everything is the same as Table \ref{comp}.}
\end{center}
\end{table*}

The proposed ORNN once again has all the best metrics except for
Wt. Acc. which is the 2nd best. 
Note that if we disregard the
ordinal labels and perform classification or regression, MAE falls off
by a large margin. Setting order penalty to 0 does not deteriorate the
performance by much, however, the percent of conflicting binary
predictions (see Section \ref{sec:multi-label}) rises from 1.4\% to
5.2\%. So adding an order penalty helps produce more interpretable
results\footnote{It is possible to increase the order penalty to
  further reduce or eliminate conflicting predictions, but we find
  that a large order penalty harms model performance.}.

\subsection{Qualitative Analysis of Predictions}
\label{sec:qualitative}

To qualitatively evaluate how well our model predicts on raw data and
observe potential patterns in the flagged samples, we obtain
predictions on the 44,105 unlabelled ads from TNABoard with the ORNN
model trained on Trafficking-10k, then we examine the samples with high
predicted likelihood to come from traffickers. Below are the top three
samples that the model considers likely:
\begin{itemize}
\item ``amazing reviewed crystal only here till fri book now please check our site for the 
services the girls provide all updates specials photos rates reviews njfantasygirls \ldots 
look who s back amazing reviewed model samantha\ldots brand new spinner jessica 
special rate today 250 hr 21 5 4 120 34b total gfe total anything goes no limits\ldots''
\item ``2 hot toght 18y o spinners 4 amazing providers today specials\ldots''
\item ``asian college girl is visiting bellevue service type escort hair color brown eyes 
brown age 23 height 5 4 body type slim cup size c cup ethnicity asian service type 
escort i am here for you settle men i am a tiny asian girl who is waiting for a gentlemen\ldots''
\end{itemize}
Some interesting patterns in the samples with high predicted
likelihood (here we only showed three) include: mentioning of multiple
names or $>1$ providers in a single ad; possibly intentional typos and
abbreviations for the sensitive words such as ``tight'' $\rightarrow$
``toght'' and ``18 year old'' $\rightarrow$ ``18y o''; keywords that
indicate traveling of the providers such as ``till fri'', ``look who s
back'', and ``visiting''; keywords that hint on the providers
potentially being underage such as ``18y o'', ``college girl'', and
``tiny''; and switching between third person and first person
narratives.

\subsection{Emoji Analysis}

The fight against human traffickers is adversarial and
dynamic. Traffickers often avoid using explicit keywords when
advertising victims, but instead use acronyms, intentional typos, and
emojis \cite{TongE2017}. Law enforcement maintains a lexicon of trafficking flags
mapping certain emojis to their potential true meanings (e.g., the
cherry emoji can indicate an underaged victim), but compiling such a
lexicon manually is expensive, requires frequent updating, and relies
on domain expertise that is hard to obtain (e.g., insider information
from traffickers or their victims). To make matters worse,
traffickers change their dictionaries over time and regularly switch
to new emojis to replace certain keywords \cite{TongE2017}. In such a dynamic and
adversarial environment, the need for a data-driven approach in
updating the existing lexicon is evident.

\begin{figure*}[htb]
	\makebox[\textwidth][c]{\includegraphics[width=1.0\textwidth]{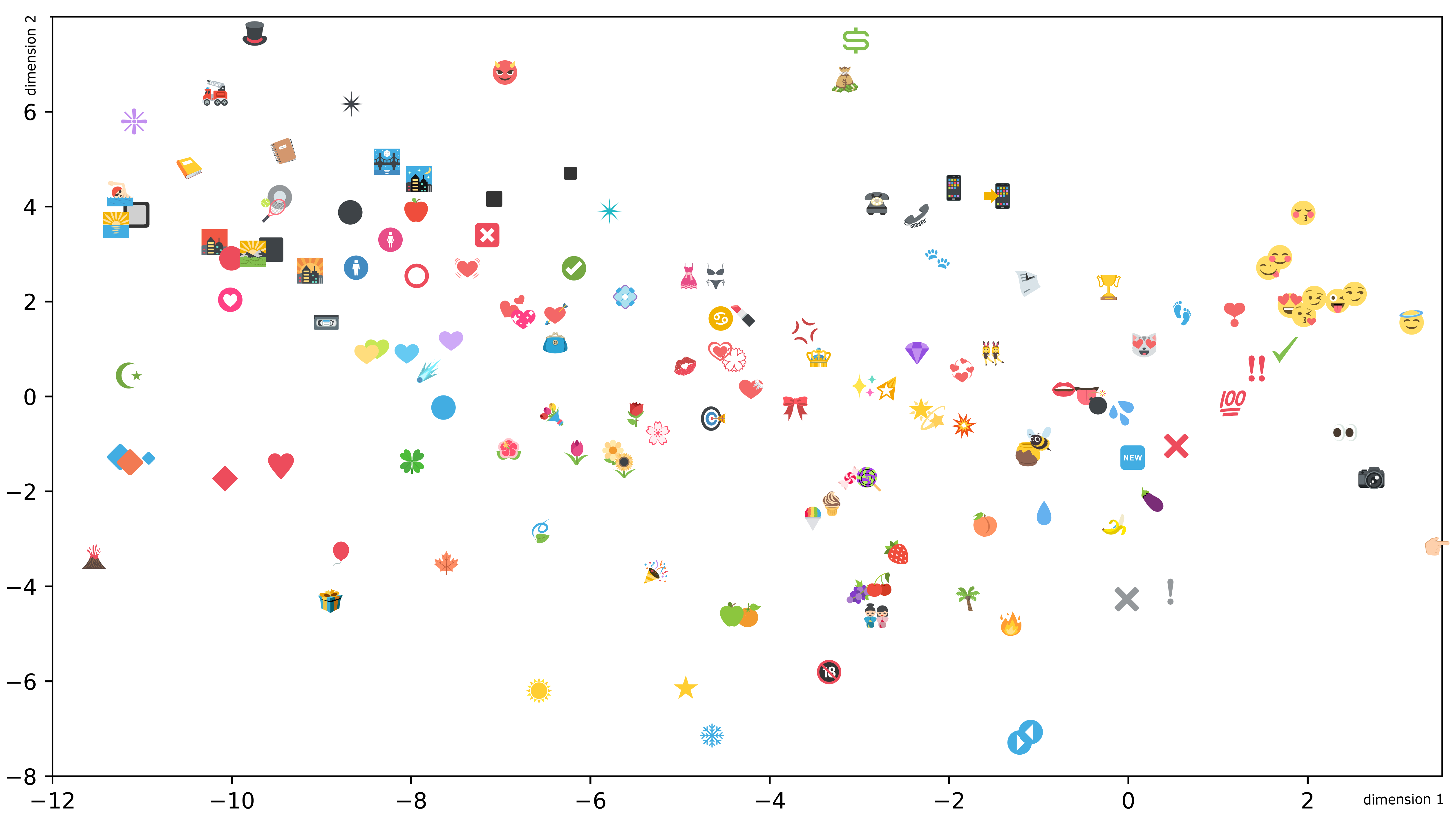}}
	\caption{\label{emoji}
          Emoji map produced by applying t-SNE to the emojis' vectors learned from escort ads 
          using skip-gram model. For visual clarity, only the emojis that appeared most frequently 
          in the escort ads we scraped are shown out of the total 968 emojis that appeared.
        }
\end{figure*}

As mentioned in Section \ref{sec:embedding}, training a skip-gram
model on a text corpus can map words (including emojis) used in
similar contexts to similar numeric vectors.  Besides using the
vectors learned from the raw escort ads to train ORNN, we can directly
visualize the vectors for the emojis to help identify their
relationships, by mapping the vectors to a 2-dimensional space using
t-SNE\footnote{t-SNE is known to produce better 2-dimensional
  visualizations than other dimension reduction techniques such as
  Principal Component Analysis, Multi-dimensional Scaling, and Local
  Linear Embedding \cite{MaatenL2008}.} \cite{MaatenL2008}
(Figure \ref{emoji}).

We can first empirically assess the quality of the emoji map by noting that similar emojis do seem 
clustered together: the smileys near the coordinate (2, 3), the flowers near (-6, -1), the heart 
shapes near (-8, 1), the phones near (-2, 4) and so on. It is worth emphasizing that the skip-gram 
model learns the vectors of these emojis based on their contexts in escort ads and not their visual 
representations, so the fact that the visually similar emojis are close to one another in the map 
suggests that the vectors have been learned as desired.

The emoji map can assist anti-trafficking experts in expanding the existing lexicon of trafficking flags. 
For example, according to the lexicon we obtained from Global Emancipation 
Network\footnote{Global Emancipation Network is a non-profit organization dedicated to combating 
human trafficking. For more information see https://www.globalemancipation.ngo.}, the cherry emoji 
and the lollipop emoji are both flags for underaged victims. Near (-3, -4) in the map, right next to these two emojis 
are the porcelain dolls emoji, the grapes emoji, the strawberry emoji, the candy emoji, the ice 
cream emojis, and maybe the 18-slash emoji, 
indicating that they are all used in similar contexts and perhaps should all be flags for 
underaged victims in the updated lexicon. 

If we re-train the skip-gram model and update the emoji map periodically on new escort ads, when 
traffickers switch to new emojis, the map can link the new emojis to the old ones, assisting 
anti-trafficking experts in expanding the lexicon of trafficking flags. 
This approach also works for acronyms and deliberate typos.

\section{Discussion}
\label{sec:dis}

Human trafficking is a form of modern day slavery that victimizes
millions of people. It has become the norm for sex traffickers to use
escort websites to openly advertise their victims.
We designed an ordinal regression neural network (ORNN) to predict the
likelihood that an escort ad comes from a trafficker, which can
drastically narrow down the set of possible leads for law
enforcement. Our ORNN achieved the state-of-the-art performance on
Trafficking-10K \cite{TongE2017}, outperforming all baseline
ordinal regression models as well as improving the classification
accuracy over the Human Trafficking Deep Network
\cite{TongE2017}. We also conducted an emoji analysis and showed
how to use word embeddings learned from raw text data to help expand the
lexicon of trafficking flags.

Since our experiments, there have been considerable advancements in
language representation models, such as BERT \cite{DevlinJ2018}.
The new language representation models can be combined with our
ordinal regression layer, replacing the skip-gram model and GF-RNN, to
potentially further improve our results.  However, our
contributions of improving the cost function for ordinal regression
neural networks, qualitatively analyzing patterns in the predicted
samples, and expanding the trafficking lexicon through a data-driven
approach are not dependent on a particular choice of language
representation model.

As for future work in trafficking detection, we can design multi-modal ordinal regression networks that 
utilize both image and text data.
But given the time and resources required to label escort ads, we may explore more unsupervised learning 
or transfer learning algorithms, such as using object 
detection \cite{RenS2015} and matching algorithms to match hotel rooms in the images.

\section*{Acknowledgments}
We thank Cara Jones and Marinus Analytics LLC for sharing the Trafficking-10K dataset. 
We thank Praveen Bodigutla for his suggestions on Natural Language Processing literature.

\appendix

\section*{Supplemental Materials}
\subsection*{Hyperparameters of the Proposed Ordinal Regression Neural Network}

\textbf{Word Embeddings}: 
speedup method: negative sampling; 
number of negative samples: 100; noise distribution: unigram distribution raised to 3/4rd; 
batch size: 16;  window size: 5; minimum word count: 5; number of epochs: 50; embedding size: 128; 
pretraining learning rate: 0.2; fine-tuning learning rate scale: 1.0.

\noindent \textbf{GF-RNN}: hidden size: 128; dropout: 0.2; number of layers: 3; gradient clipping norm: 0.25; 
L2 penalty: 0.00001; learning rate decay factor: 2.0; learning rate decay patience: 3; 
early stop patience: 9; batch size: 200; 
output layer type: mean-pooling; minimum word count: 5; maximum input length: 120.

\noindent \textbf{Multi-labeled Logistic Regression Layer}: task weight scheme: uniform; conflict penalty: 0.5.

\subsection*{Access to the Source Materials}
The fight against human trafficking is adversarial, hence the access to the source materials in anti-trafficking research is typically not available to the general public by choice, but granted to researchers and law enforcement individually upon request.

\noindent \textbf{Source code}: 
https://gitlab.com/BlazingBlade/TrafficKill

\noindent \textbf{Trafficking-10k}: cara@marinusanalytics.com

\noindent \textbf{Trafficking lexicon}: sherrie@globalemancipation.ngo

\fontsize{9.8pt}{10.8pt} \selectfont
\nocite{*}
\bibliographystyle{aaai}
\bibliography{myBib}

\end{document}